\documentclass{article}

\usepackage{arxiv}
\usepackage[utf8]{inputenc}
\usepackage[T1]{fontenc}
\usepackage[hidelinks]{hyperref}
\usepackage{url}
\usepackage{booktabs}
\usepackage{amsfonts}
\usepackage{nicefrac}
\usepackage{microtype}
\usepackage{graphicx}
\usepackage[numbers]{natbib}
\usepackage{doi}
\usepackage{amsmath}

\title{OUTPATIENT APPOINTMENT SCHEDULING OPTIMIZATION WITH A GENETIC ALGORITHM APPROACH}

\date{February 25, 2026}

\author{ \href{https://orcid.org/0009-0006-3252-7343}{\includegraphics[scale=0.06]{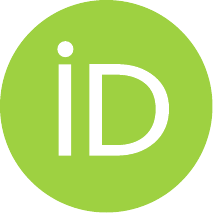}\hspace{1mm}Ana C. Rodrigues}\\
	Crab Technologies, Lda\\
	Rua do Poço, 28, 4900-519, \\
        Viana do Castelo, Portugal \\
	\texttt{ana.rodrigues@crabtech.io} \\
        \And
	{Rui Rego} \\
	Crab Technologies, Lda\\
	Rua do Poço, 28, 4900-519,\\
        Viana do Castelo, Portugal \\
	\texttt{rui.rego@crabtech.io} \\
}

\renewcommand{\headeright}{}
\renewcommand{\undertitle}{Technical Report}
\hypersetup{
pdftitle={OUTPATIENT APPOINTMENT SCHEDULING OPTIMIZATION WITH A GENETIC ALGORITHM APPROACH},
pdfsubject={q-bio.NC, q-bio.QM},
pdfauthor={Ana C. Rodrigues, Rui Rego},
pdfkeywords={Genetic Algorithm,Healthcare,Medical Appointment Scheduling},
}

\begin{document}
\maketitle

\begin{abstract}
	The optimization of complex medical appointment scheduling remains a significant operational challenge in multi-center healthcare environments, where clinical safety protocols and patient logistics must be reconciled. This study proposes and evaluates a Genetic Algorithm (GA) framework designed to automate the scheduling of multiple medical acts while adhering to rigorous inter-procedural incompatibility rules. Using a synthetic dataset encompassing 50 medical acts across four healthcare facilities, we compared two GA variants — Pre-Ordered and Unordered — against deterministic First-Come, First-Served (FCFS) and Random Choice baselines. Our results demonstrate that the GA framework achieved a 100\% constraint fulfillment rate, effectively resolving temporal overlaps and clinical incompatibilities that the FCFS baseline failed to address in 60\% and 40\% of cases, respectively. Furthermore, the GA variants demonstrated statistically significant improvements ($p < 0.001$) in patient-centric metrics, achieving an Idle Time Ratio (ITR) frequently below 0.4 and reducing inter-healthcenter trips. While the GA (Ordered) variant provided a superior initial search locus, both evolutionary models converged to comparable global optima by the 100th generation. These findings suggest that transitioning from manual, human-mediated scheduling to an automated metaheuristic approach enhances clinical integrity, reduces administrative overhead, and significantly improves the patient experience by minimizing wait times and logistical burdens.
\end{abstract}

% keywords can be removed

\section{Introduction}

The coordination of multiple medical examinations within a single patient journey represents a significant operational challenge in modern healthcare systems. This complexity arises from the necessity to synchronize diverse resources—including specialized practitioners, specific diagnostic equipment, and physical room availability—while accounting for high variability in service durations and patient arrivals \cite{ahmadi2017outpatient}. Furthermore, clinical interdependencies, such as inter-procedural incompatibilities and mandatory recovery gaps, introduce rigid temporal constraints that must be strictly satisfied to ensure diagnostic integrity and patient safety \cite{marynissen2019literature}. Manual scheduling in these environments is often inefficient, time-consuming, and prone to human error, frequently resulting in fragmented patient journeys characterized by excessive wait times, redundant travels, and unsatisfied patients \cite{cayirli2003outpatient,vissers2001framework}. 

Genetic Algorithms (GAs) offer a robust solution to these challenges by providing a stochastic global search heuristic capable of navigating complex, non-linear search spaces \cite{david1989goldberg}. Categorized as an NP-hard optimization problem, medical appointment scheduling exceeds the practical capabilities of exact mathematical solvers as the number of variables and constraints increases \cite{garey2002computers,pinedo1992scheduling}. GAs address this by simulating evolutionary processes to iteratively evolve a population of candidate solutions towards a global optimum \cite{holland1992adaptation}.

Recent literature has explored the application of evolutionary computation in healthcare scheduling. Squires et al \cite{squires2022novel}, presented a List Scheduling Wildcard Tournament Genetic Algorithm to optimise the operational efficiency of scheduling repetitive Transcranial Magnetic Stimulation appointments. Lin et al \cite{lin2020hybrid}, presented a hybrid genetic algorithm for operating room scheduling, designed to maximize the utilization of operating rooms, minimize the overtime-operating cost, and minimize wasting cost for unused time. Gombás et al \cite{gombas2025outpatient}, analysed the impact of different mutation operators in the performance of a genetic algorithm developed for outpatient scheduling.

Despite this, there still remains a need for research on scheduling algorithms that prioritize the patient-centric experience by minimizing geographical and temporal fragmentation of scheduled appointments. This research seeks to address the following question: To what extent can a Genetic Algorithm optimize the efficiency of scheduling multi-appointment patient journeys while strictly adhering to complex clinical constraints?

\section{Methodology}

\subsection{Synthetic Data Generation}

To model the high complexity of medical data, we started by defining 5 specialities (Radiology, Cardiology, Dermatology, General Practice, and Gastroenterology) and generating 10 exams for each.

Inter-procedural constraints were modelled via incompatibility rules. These rules dictate the required temporal separation between specific exam pairs to prevent physiological or diagnostic interference (e.g., the influence of a stress test ECG on a subsequent resting ECG). Incompatibilities were defined by an ordered pair $(E_1, E_2)$, a constraint logic (selected from {"before", "after", "both"}), and a mandatory gap duration (selected from {30, 60, 1440} minutes). Both logic and duration were assigned via uniform probability distributions. A total of 15 stochastic incompatibility rules were implemented.

The temporal search space spanned 30 days across four healthcare facilities, each containing three examination rooms. Operating hours were set from 09:00 to 21:00. Slot generation followed a stochastic iterative process: slot duration was assigned from {15, 30, 45, 60, 90} minutes via uniform distribution; practitioner was randomly assigned from a pool of four clinicians; a random examination type was assigned to each generated slot until the daily temporal limit was reached.

\subsection{The Genetic Algorithm}

Genetic algorithms (GAs) are stochastic global search heuristics inspired by the principles of Neo-Darwinian evolution. They operate on a population of candidate solutions (individuals), iteratively applying natural selection inspired operations to navigate complex, non-linear search spaces.

\subsubsection{Defining the search space}

The total pool of available time slots for each request is retrieved from the database. These are filtered based on medical acts requested, start date for the search, and user-defined preferences regarding healthcare centres and practitioners.

\subsubsection{Creating the initial population}

The process begins with the definition of an “individual”. Each individual is a binary array representing a potential solution. A population size of $N=100$ was maintained in this work, to ensure broad coverage of the initial search space.

For each medical act requested, a binary array is generated, each element corresponding to one available slot. Only one slot is selected (set to 1) for each medical act. The final individual corresponds to the concatenation of the array for each act. 

Two approaches were tested for the creation of the population. A baseline approach consisted in randomly selecting one time slot for each of the requested acts. While a “pre-ordered” approach consisted in randomly selecting one time slot for each of the requested acts while keeping to an optimal act order calculated previously. This optimal order was calculated based on the exam incompatibilities data.

Two initialization strategies were evaluated:

\begin{itemize}
    \item GA Unordered: Stochastic selection of one slot per requested act.
    \item GA Ordered: Stochastic selection restricted to an optimal sequence pre-calculated based on known exam incompatibilities to minimize initial constraint violations.
\end{itemize}

\subsubsection{Fitness evaluation}

Each individual within the population is subjected to a fitness function, f(x). This objective function quantifies the optimality of a candidate solution relative to the defined constraints. The fitness function is defined as:
$$F(x) = \frac{1}{C_{base} - \sum{C_{penalties}}}$$
Where:
\begin{itemize}
    \item $C_{base}$ = 1 (to prevent division by zero)
    \item $C_{penalties}$:
    \subitem $+1000$ if number of selected slots $\neq$ number of requested acts.
    \subitem $+1000$ for temporal overlaps or incompatibility breaches.
    \subitem $+100 \times N_{trips}$ (a trip is defined by a facility change or a temporal gap $> 2$h).
    \subitem $+600$ if the gap between disparate trips is $< 3$h.
    \subitem $+(\text{Total Wait Time} / 10) + (\text{Days from start\_date to first slot})$.
\end{itemize}

\subsubsection{Selection}

Selection mechanisms simulate "survival of the fittest" by filtering the population. In this work, a tournament approach was applied, where subgroups of $k=7$ individuals were randomly sampled, and the superior individual was selected. This was repeated $N=100$ times (population size).

\subsubsection{Crossover}

Crossover facilitates the exchange of genetic material between two "parent" individuals to produce "children" individuals. This work applied single-point crossover, where a locus (randomly selected index) is chosen at which the genetic strings (binary arrays) are swapped.

\subsubsection{Mutation}

A mutation rate of 10\% was considered, meaning that, for 10\% of the children, the selected exam slot will be randomly reassigned to one of the available slots. Mutation introduces random perturbations into the population, maintaining genetic diversity and preventing the algorithm from becoming trapped in local optima.
The best performing individual of the previous generation was included into the new generation, and the process was repeated for 200 generations.

\subsection{Baseline algorithms}

The GA performance was benchmarked against two standard scheduling heuristics:

\begin{itemize}
    \item First-Come, First-Served (FCFS): A deterministic approach that assigns the earliest available chronological slot for each required exam, regardless of total journey duration or facility location.
    \item Random Choice: A stochastic approach where one valid slot is selected at random for each required act, serving as a control to measure the optimization gain of the GA.
\end{itemize}

\subsection{Performance Metrics}

To evaluate the efficacy of the GA and baseline approaches, four primary metrics were quantified:

\begin{itemize}
    \item Idle Time Ratio (ITR): Quantifies the temporal efficiency of a patient’s scheduled journey. It is defined as the proportion of unutilized time between scheduled appointments within the total duration of the patient’s journey. Calculated as:
    $$ITR = \frac{\sum (T_{start, i+1} - T_{end, i})}{T_{end, n} - T_{start, 1}}$$
where $T_{start, 1}$ is the commencement of the first scheduled slot and $T_{end, n}$ is the conclusion of the final slot. An ITR approaching 0 indicates maximum temporal efficiency.
    \item Inter-facility Displacement: The total number of required facility changes (trips) per solution. This metric serves as a proxy for patient logistical burden; a higher count indicates a sub-optimal solution.
    \item Constraint Violation Rate: The frequency of constraints violation within the final solution. Three constraints were considered: 1) outputted timeslots must not temporally overlap; 2) exam-pair incompatibility rules and mandatory gaps must be strictly respected; 3) a minimum threshold of $\ge 3$ hours must be maintained between appointments requiring inter-facility displacement to ensure logistical viability.
\end{itemize}

\section{Results}

\begin{figure}[h!]
\centering
\includegraphics[width=14cm]{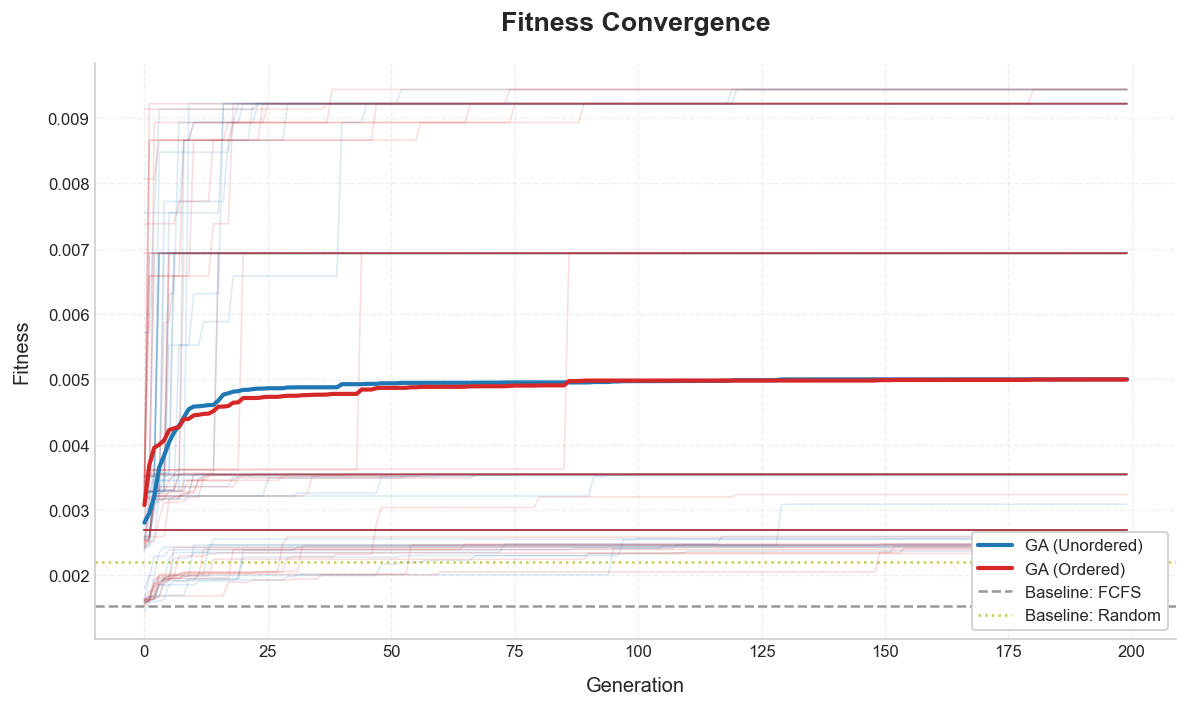}
\caption{\label{fitness}Mean Fitness Convergence Profiles across 200 Generations. This plot illustrates the evolutionary progression of the GA (Ordered) and GA (Unordered) variants compared to the FCFS and Random Choice baselines.}
\end{figure}

The fitness convergence profiles (Figure \ref{fitness}) indicate a significant performance delta between the evolutionary algorithms and the baseline heuristics. The Genetic Algorithm (GA) variants exhibited rapid optimization, surpassing both Random Choice and First-Come, First-Served (FCFS) baselines within the first 10 generations.

Both GA variants exhibit rapid fitness gains in the initial evolutionary phase, reaching a position with minimal fluctuation after 50 generations and remaining stable after 100 generations, indicating convergence toward a global optimum, significantly exceeding the performance established by the baseline algorithms.

The GA (Ordered) approach initiates with a higher mean fitness compared to the GA (Unordered) variant, suggesting that incorporating domain-specific constraints (exam incompatibilities) into the initial population generation provides a superior starting locus in the search space.

\begin{figure}[h!]
\centering
\includegraphics[width=14cm]{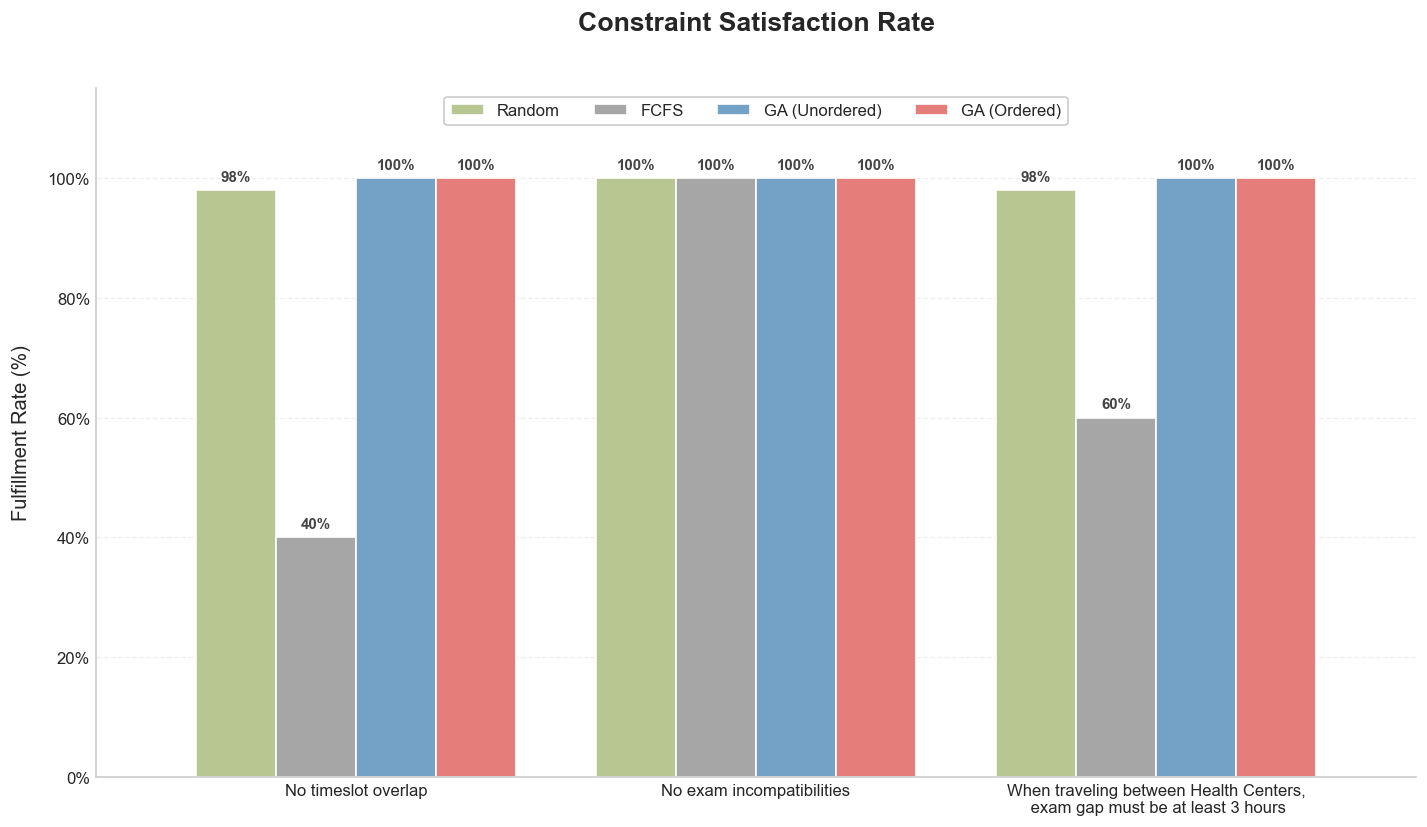}
\caption{\label{constraints}Constraint Fulfillment Rates. A categorical breakdown of the percentage of solutions satisfying the three primary scheduling constraints: 1) temporal non-overlap, 2) clinical compatibility (exam gaps), and 3) transit feasibility ($\ge 3$h).}
\end{figure}

Both GA variants achieved a 100\% fulfillment rate across all three primary constraints (Figure \ref{constraints}). The FCFS baseline failed to resolve timeslot overlaps in 60\% of cases and violating travel gap requirements in 40\% of instances. While the Random Choice model maintained high clinical compatibility, it failed to guarantee 100\% reliability in overlap and travel constraints.

\begin{figure}[h!]
\centering
\includegraphics[width=14cm]{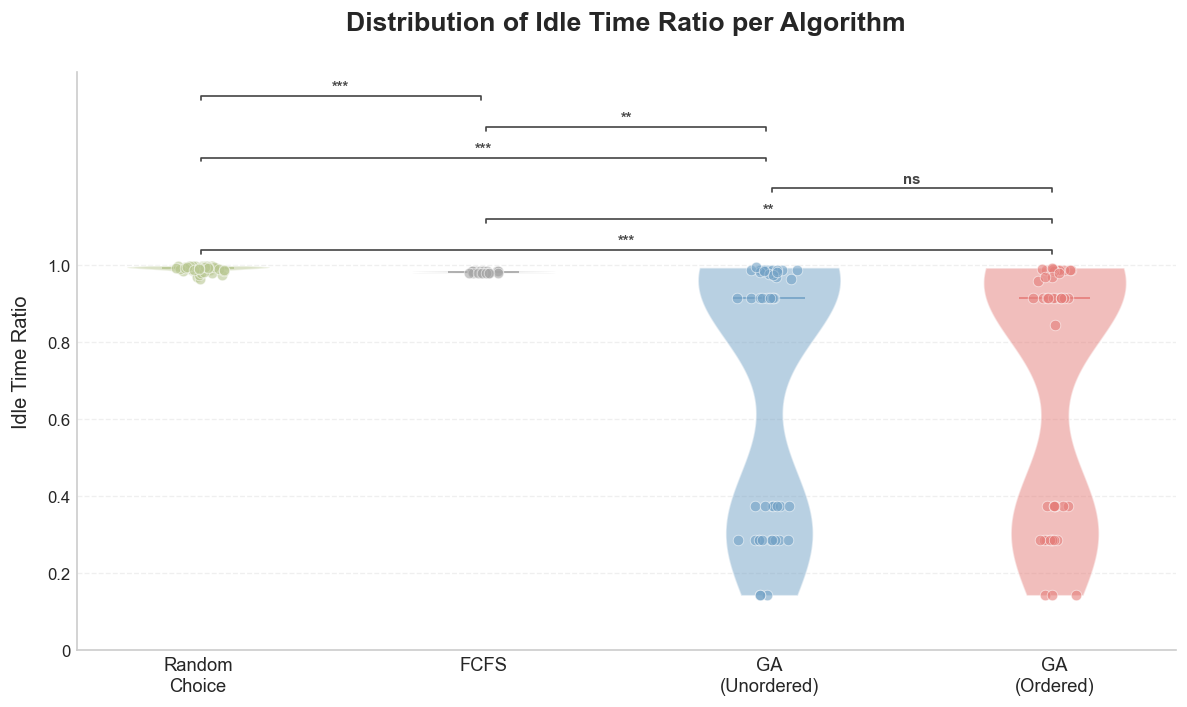}
\caption{\label{itr}Statistical Distribution of the Idle Time Ratio (ITR). Violin plots representing the density and distribution of the ITR for each tested algorithm. Statistical comparison performed with a Mann-Whitney U test. ns: $p > 0.05$ (not significant); *: $p < 0.05$; **: $p < 0.01$; ***: $p < 0.001$}
\end{figure}

Both GA variants achieved statistically significant reductions in ITR (Figure \ref{itr}) compared to baselines ($p < 0.001$). While the Random and FCFS methods clustered near an ITR of 1.0, the GA variants produced solutions, frequently achieving ratios below 0.4. No statistically significant difference was observed between the Unordered and Ordered GA variants regarding final ITR.

\begin{figure}[h!]
\centering
\includegraphics[width=14cm]{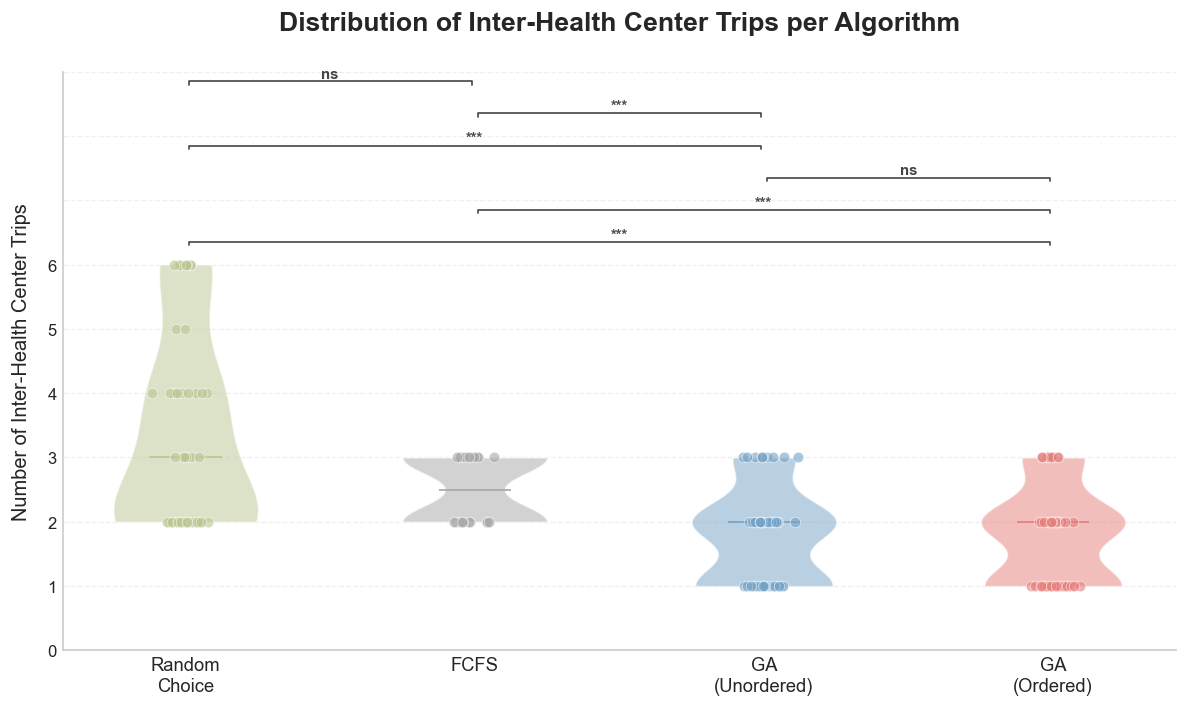}
\caption{\label{trip}Frequency of Inter-Health Center Transitions. Distribution of the total number of required facility changes per patient journey. Statistical comparison performed with a Mann-Whitney U test. ns: $p > 0.05$ (not significant); *: $p < 0.05$; **: $p < 0.01$; ***: $p < 0.001$}
\end{figure}

Similarly, the GA variants significantly minimized patient mobility requirements (Figure \ref{trip}), with a median of 2 trips per solution. This outperformed the FCFS baseline ($p < 0.001$) and the Random Choice model, which exhibited a median of 2.5 and 3 trips, respectively.

\section{Discussion}

The Genetic Algorithm (GA) approach presented in this work offers a robust computational framework for resolving the multi-objective optimization problem inherent in complex medical appointment scheduling. Unlike manual scheduling workflows, this automated framework provides a scalable and error-resistant mechanism for enhancing operational efficiency in multi-center healthcare environments.

The failure of the First-Come, First-Served (FCFS) approach, simulating the typical human workflow in the scheduling of medical appointments, indicates that greedy, deterministic logic is insufficient for non-linear scheduling spaces. While FCFS prioritizes immediate availability, it fails to account for the downstream effects of inter-procedural incompatibilities and geographical constraints, leading to a high Idle Time Ratio (ITR).

Conversely, the GA approaches were able to maintain a 100\% fulfillment rate across all constraints. The high penalty associated with overlaps and incompatibilities (cost +1000) effectively pruned non-viable candidates from the gene pool, ensuring that final solutions were clinically and logistically feasible. The statistical superiority ($p < 0.001$) of the GA in reducing the ITR and inter-facility displacement further proves its success in navigating the search space to find optimal solutions.

The comparative data between GA (Ordered) and GA (Unordered) suggests that while initialization influences the initial search locus, the evolutionary process is robust enough to reach comparable global optima regardless of the starting state. The higher initial mean fitness of the GA (Ordered) variant confirms that pre-calculating optimal sequences based on exam incompatibilities reduces the computational overhead required to eliminate early-stage constraint violations. However, the convergence of both models to a similar plateau by generation 100 implies that for the defined search space, the selection pressure and genetic diversity were sufficient to overcome the lack of initial ordering. Despite this, the impact of initial ordering might be an interesting strategy in contexts where algorithmic response time is a concern, allowing for a higher-fitness solution, in fewer generations.

The implementation of the proposed framework in a clinical setting offers several operational advantages over traditional manual scheduling workflows. Currently, multi-exam optimization, encompassing clinical incompatibilities, geographical logistics, and temporal density, relies on manual scheduling performed by front-office and contact center personnel, which is inherently susceptible to human error. Transitioning to an algorithmic model automates these high-cognitive-load tasks, significantly reducing the human resource expenditure required for complex patient journeys, and ensuring compliance with the defined safety constraints. Furthermore, traditional scheduling often requires patients to initiate contact to resolve complex bookings, leading to potential attrition among patients who prefer streamlined digital interactions. Automating the selection process enables the proactive suggestion of optimized slot clusters, thereby minimizing patient loss and increasing total lead conversion. Finally, the proposed approach directly enhances patient satisfaction by minimizing wait time between appointments and the frequency of inter-facility trips.

\section{Conclusion}

This study demonstrates that the application of a Genetic Algorithm (GA) provides a superior alternative to deterministic and stochastic baseline heuristics for the complex task of medical appointment scheduling. The transition from manual, error-prone scheduling to this automated metaheuristic framework enhances clinical integrity, reduces administrative overhead, and maximizes patient satisfaction through superior booking optimization.

\section{Acknowledgements}

This work received funding from Crab Technologies, Lda \cite{crab}.

\bibliographystyle{unsrtnat}
\bibliography{template}

\end{document}